%% file: 0_root.tex
\documentclass[letterpaper, 10 pt, conference]{ieeeconf}  

\IEEEoverridecommandlockouts                              

\usepackage{graphicx} 
\usepackage{amsmath} 
\usepackage{amssymb}  
\usepackage{cite}
\usepackage{float}

\usepackage{nidanfloat}

\title{\LARGE \bf
Acoustic Communication and Sensing for \\ Inflatable Modular Soft Robots
}

\author{Daniel S. Drew$^{1}$, Matthew Devlin$^{2}$, Elliot Hawkes$^{2}$, and Sean Follmer$^{1}$%
\thanks{$^{1}$Daniel S. Drew and Sean Follmer are with the Department of Mechanical Engineering, Stanford University, Stanford, CA, USA.
        {\tt\small dsdrew@stanford.edu}}%
\thanks{$^{2}$Matthew Devlin and Elliot Hawkes are with the Department of Mechanical Engineering, University of California Santa Barbara, Santa Barbara, CA, USA.}
      
}

\begin{document}

\maketitle



\thispagestyle{empty}
\pagestyle{empty}

\begin{figure*}[!b]
    \centering
    \includegraphics[
    height=1.5in]{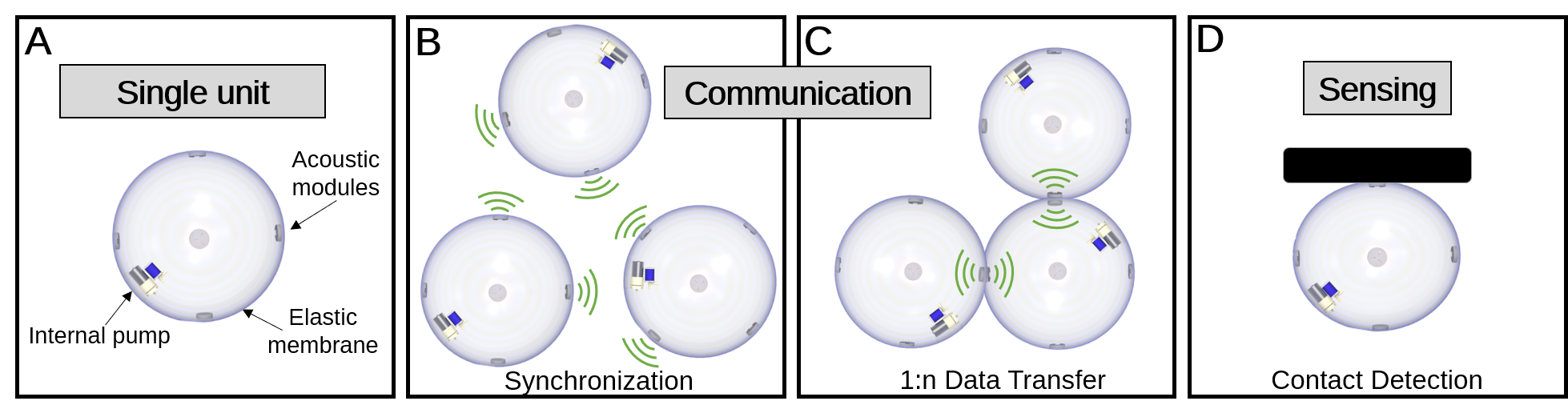}
    \caption{Inflatable soft modular robots. A) Each robot unit comprises a latex membrane, an internal pump and release valve, and ``acoustic modules,'' consisting of piezoelectric transducers attached to magnetic connectors, distributed over their internal surface. These acoustic modules are both able to overcome challenges associated with instrumenting soft, high extension ratio robots as well as being scalable and efficient enough to enable modular, multi-robot systems. The proposed architecture allows (B) communication at a distance for synchronization, (C) directional neighbor-to-neighbor data transfer, and (D) external contact sensing.}
    \label{fig:teaser}
    \vspace{-4mm}
\end{figure*}
\begin{abstract}

\input{1_abstract}

\end{abstract}

\input{2_introduction}
\input{3_relatedwork}
\input{4_implementation}
\input{5_behaviors}
\input{6_futurework}



\section*{ACKNOWLEDGMENT}
This work was supported in part by the Intelligence Community Postdoctoral Research
Fellowship Program, administered by the Oak Ridge Institute for Science and
Education through an Interagency Agreement between the U.S. DoE and ODNI

\bibliographystyle{IEEEtran}
\bibliography{references}

\end{document}

%% file: 1_abstract.tex
Modular soft robots combine the strengths of two traditionally separate areas of robotics. As modular robots, they can show robustness to individual failure and reconfigurability; as soft robots, they can deform and undergo large shape changes in order to adapt to their environment, and have inherent human safety. 
However, for sensing and communication these robots also combine the challenges of both: they require solutions that are scalable (low cost and complexity) and efficient (low power) to enable collectives of large numbers of robots, and these solutions must also be able to interface with the high extension ratio elastic bodies of soft robots. In this work, we seek to address these challenges using acoustic signals produced by piezoelectric surface transducers that are cheap, simple, and low power, and that not only integrate with but also leverage the elastic robot skins for signal transmission. Importantly, to further increase scalability, the transducers exhibit multi-functionality made possible by a relatively flat frequency response across the audible and ultrasonic ranges. With minimal hardware, they enable directional contact-based communication, audible-range communication at a distance, and exteroceptive sensing. 
We demonstrate a subset of the decentralized collective behaviors these functions make possible with multi-robot hardware implementations. The use of acoustic waves in this domain is shown to provide distinct advantages over existing solutions.

%% file: 2_introduction.tex
\section{Introduction}

Modular robots overcome individual platform limitations by physically connecting and reconfiguring in order to tailor their system-level capabilities to their application and environment~\cite{brunete_current_2017}.
At the same time, soft, shape-changing robots have distinct advantages over rigid-bodied robots, including passive adaptation to their environment through structural compliance, inherent safety for human-robot interaction tasks, and the ability to exert relatively large forces and undergo relatively large strains with low-cost actuators~\cite{zhang_modular_2020}. 
Modular soft robots, which take inspiration from biological collectives (as ``cellular robots''~\cite{yu_morpho_2008}), combine these advantages in order to perform useful behaviors emergent from interactions between relatively simple individual units. 
A major barrier to progress, however, is the fact that these robots also combine the challenges of these two realms. 
For example, a significant challenge in the design of modular robots meant to be deployed in large collectives is balancing individual platform size, complexity, and cost with the architecture and functionality of the conjoined system. 
The design of multi-functional components, which can adequately fulfill the function of multiple robotic subsystems without requiring additional hardware, is a potential solution.
The soft, extensible structure of a modular soft robot compounds the challenge by placing additional constraints on the possible implementations, which must be both robust to high extension ratios as well as able to be coupled to elastic surfaces.

Many modular and swarm robots have sought to address the challenge of scalable inter-agent communication and sensing via infrared (IR) optical transmission~\cite{brunete_current_2017}. This relatively low range and line-of-sight constrained method may be supplemented by wider area radio-frequency networking~\cite{seo_modular_2019}.
In contrast, in nature the use of \textit{acoustic} signals is ubiquitous, including among the social insects which inspire many designers of modular and swarm robots~\cite{cocroft_public_2011}. These acoustic signals include substrate-borne vibrations, audible sound, and vibrations shared through direct body contact~\cite{hill_how_2009,holldobler1999multimodal}. 
Inspired by the way that existing organisms use passive mechanical body structures to efficiently produce, receive, and transmit acoustic signals -- from the audible range of the cricket~\cite{montealegre-z_sound_2011} to the ultrasonic range of the moth~\cite{spangler1984ultrasonic} -- the same pre-tensioned elastic membranes that make soft robots so difficult to instrument for sensing and communication make them particularly attractive for \textit{multi-functional acoustics-based components.}


Existing acoustic transducers are well-suited for acting as multi-functional components due, in part, to their ability to be operated across a wide spectrum. The Huygens-Fresnel principle dictates that the directivity of a wave corresponds to the size of the source relative to the wavelength. In practice this change in directivity is beneficial for applications like ultrasonic obstacle detection~\cite{bank_novel_2002}, where it limits the field-of-view of the transducer and focuses the signal just as a lens does for an infrared source, and is a challenge for designers of speakers with desirable ``dispersion patterns.'' 
In addition to this variable directivity, the attenuation of acoustic waves in air is proportional to the wave frequency; the absorption coefficient of air increases approximately 30dB from 1kHz to 20kHz~\cite{bass_atmospheric_1995}. 
The relatively flat frequency response (up to about 20kHz) of the simple commodity piezoelectric disc transducers used in this work therefore means that they can be operated with variable attenuation and directionality depending on desired function. 
\begin{figure}
    \centering
    \includegraphics[width=.8\columnwidth]{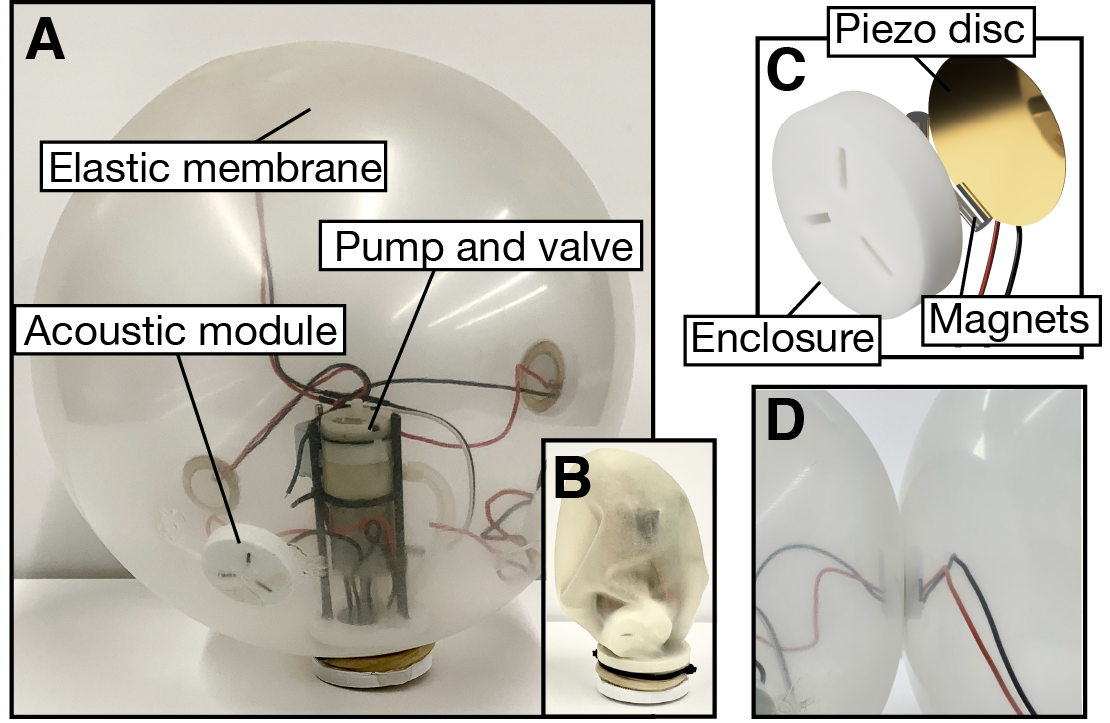}
    \caption{A) An inflated robot with acoustic modules dispersed on its interior surface. B) The robot, when fully deflated, is roughly the same size as the pump it contains. C) The acoustic modules are composed of 3D printed enclosures containing three diametrically polarized cylindrical magnets with an affixed piezoelectric transducer. D) Magnetic connection between two adjacent inflated robots, made through their acoustic modules. }
    \label{fig:hardware}
    \vspace{-4mm}
\end{figure}

The contribution of this work is a communication and sensing modality (Fig.~\ref{fig:teaser}) based on surface-distributed ``acoustic modules,'' which use piezoelectric transducers to both send and receive acoustic waves across the audible to ultrasonic spectra, implemented on modular soft robots with high extension ratios (Fig.~\ref{fig:hardware}). The modules are scalable (i.e, of minimal cost and complexity), efficient (i.e., each module consumes 60mW, compared to the 160mW IR emitter of the Kilobot~\cite{rubenstein_kilobot_2012}), and helps to perform multiple core robotic functions. They not only integrate simply with elastic skins through surface attachment, they also take advantage of the structure itself as a transmission medium that is robust to large shape change. Together, this makes our solution cost effective, capable, and versatile compared to other options for shape-changing modular soft robots. After discussing related work in Section \ref{related}, we describe the technical implementation and results of testing in Section \ref{implementation}, showing core collective functions like communication between robots, synchronization at a distance, and sensing of external stimuli. In Section \ref{behaviors}, we demonstrate two enabled collaborative behaviors in a group of three robots, including synchronized lifting and a decentralized inchworm-based gait.

%% file: 3_relatedwork.tex
\section{Related Work} \label{related}

The most directly relevant related work includes other modular and swarm robots that use individual subsystems or components for multiple functions and other examples of acoustic sensing and communication in multi-robot systems.


\subsection{Multifunctional Hardware for Multi-robot Systems}

The Linbot soft modular platform~\cite{mckenzie_linbots_2018} is the most directly related to this work. It uses a voice coil for actuation, sensing, and communication, taking advantage of the wide frequency response in a similar manner to how we use our piezoelectric transducers. A Hall-effect sensor is used for proprioception through sensing of the voice coil position, electromagnetic coupling between neighboring Linbots allows for omnidirectional communication, and audible range waves can be produced for external communication. To accomplish this they rely on the rigid connections between the actuator core and the extents of the soft shell, only operating with shape changes of up to approximately 30$\%$ on their principle axis. In contrast, our robots undergo maximum volume changes of close to 1000$\%$, and the exterior surfaces do not remain in contact with the primary actuator.

Swarm platforms are relevant in this context because they are also motivated by finding low complexity and cost, scalable solutions~\cite{brambilla_swarm_2013}. The Kilobot~\cite{rubenstein_kilobot_2012} platform uses an IR transmitter and receiver on its underside to both communicate with and detect the distance of neighbors, using only one pair for both functions but doing so only omnidirectionally and only up to about 10cm away. The Open E-Puck platform~\cite{gutierrez_open_2009} uses a set of 12 pairs of radially arranged IR transmitters and receivers to perform inter-robot communication as well as range and bearing measurements, which allows it to send and receive signals from specific directions. Our acoustic solution adds the additional functionality of long-range ($>$1m) communication with no line-of-sight requirements, as well as contact/deformation sensing, while only requiring a single transducer instead of an emitter/receiver pair.


\subsection{Multi-robot Acoustic Sensing and Communication}

A common use of acoustic waves in multi-robot systems is for ultrasonic range estimation. The relatively slow speed of sound lessens signal processing constraints relative to radio frequency solutions (e.g., RSSI) by enabling direct time of flight measurements, making it a useful supplement to improve robustness of distance estimation~\cite{girod_robust_2001}. Relative positioning of multi-robot systems using ultrasonic ranging at distances up to seven meters has been demonstrated with absolute average error of only 8mm~\cite{rivard_ultrasonic_2008}.

As opposed to sensing, acoustic communication between autonomous robots is a relatively underexplored area. An exception is in the realm of autonomous underwater vehicles, which are driven towards acoustic modes by the high electromagnetic absorption of seawater~\cite{chappell_acoustic_1994, bahr_cooperative_2009}. Audible range communication has been noted as a potentially useful supplement to radio frequency networking for land-based multi-robot systems due to the fact that the relatively strong environmental attenuation of acoustic waves can encode environmental information~\cite{karimian_sounds_2006}. In this work, we take this idea further by using the soft pressurized structure of the robot itself as the information-encoding transmission environment.

Outside of the robotics domain, acoustic communication has been shown between pressurized mylar balloons that act as amplifiers and speakers when actuated by piezoelectric transducers~\cite{paradiso_interactive_1996}, which served as an inspiration for the communication-at-a-distance in this work.




%% file: 4_implementation.tex
\section{Implementation and Results} \label{implementation}

\subsection{System Hardware}
\begin{figure}
    \centering
    \includegraphics[width=0.8\columnwidth]{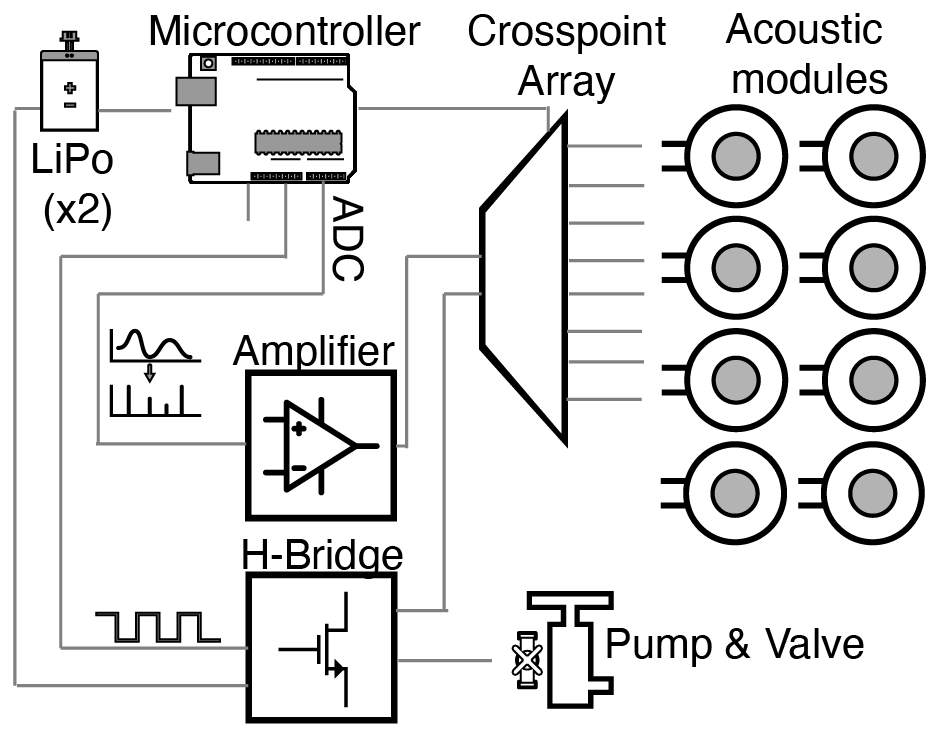}
    \caption{A system block diagram illustrating how the analog switch array is used to dynamically connect the piezoelectric actuators to either the preamplifier or to the motor driver depending on desired function.}
    \label{fig:systemblock}
    \vspace{-4mm}
\end{figure}


The inflatable robot units (Fig.~\ref{fig:hardware}A) are based on our co-authors' recent prior work~\cite{devlin_untethered_nodate} demonstrating untethered cellular robots. Each is composed of a 45cm maximum diameter, 0.4mm thick latex membrane enclosing a DC air pump, solenoid-controlled valve, and up to \textit{N} (\textit{N} = 8 given the analog switch array used in this work) acoustic modules distributed across the membrane. Each acoustic module represents a sensing, communication, and connection point for the robots to interact with their environment and each other. There is an inherent tradeoff between the increased functionality (e.g., in terms of sensing resolution) and the increased complexity for each additional acoustic module which bears future investigation.


The acoustic modules (Fig.~\ref{fig:hardware}C) comprise FDM 3D printed, cylindrical enclosures (30mm diameter, 7mm thickness, PLA) with 60 degree radially arrayed slots for diametrically polarized cylindrical magnets (3.2mm diameter, 6.4mm height). The magnet housings are slightly over-sized, allowing the magnets to reorient when connectors are drawn together, making them ``genderless.'' We rely upon passive reorientation of the units for alignment, although notably vibrations can be transmitted with sufficient signal-to-noise ratio through even imperfectly aligned modules. The piezoelectric transducers (27mm diameter brass plate with 20mm diameter ceramic piezo, 0.5mm thickness) are standard contact microphones, fixed into the printed enclosures with 3M 300LSE double sided adhesive tape. The acoustic modules are each fixed to the inside of the latex skin with the same tape.

A Teensy4, which contains a 600MHz Cortex M7 microprocessor, is sufficient for the software-defined radio architecture of the acoustic communication. To minimize cost and complexity the piezoelectric transducers are connected through an 8:16 analog switch array (MT8816) to a single full H-bridge dual-channel motor driver (TB6612FNG) and a single audio amplifier (MAX9814, includes preamplifier, variable gain stage, and output amplifier) with 60dB gain. A block diagram of the system is shown in Fig.~\ref{fig:systemblock}. Each piezoelectric transducer consumes approximately 60mW during full-duty cycle operation between 3kHz and 20kHz (10mA at 6V). The poor impedance matching between the piezoelectric transducer and the amplifier, which is designed for standard electret condenser microphones, creates a high pass filter around 2kHz.
Although here we show units with electronics and power located externally to the robot membrane, prior work shows that the required electronics, battery pack, and charging circuit can be incorporated into the latex membranes inside a 3D printed enclosure~\cite{devlin_untethered_nodate}.


\subsection{Contact-based Communication}

Swarm and modular robot systems designed for large agent counts typically rely heavily on local communication as a way to overcome challenges with scaling of radio-based networks~\cite{brambilla_swarm_2013}.
For modular robots the connection points represent natural avenues for information transfer, such as through direct electrical connections~\cite{gilpin_robot_2010}. 
Methods that do not rely on mechanically flush or material-specific connections, like IR transmit/receive pairs built into the faces of the connectors~\cite{gilpin_miche_2008}, are more suitable for deformable surfaces.
In our robots, the piezoelectric transducers in the rigid enclosure of the magnetic connectors can transfer information in the form of shared vibration through even imperfect contact made between connectors; the received signal amplitude for a 18kHz tone decreases from its full value when all three magnets are aligned, $N_{contacts}=3$, by about 35$\%$ for $N$=2 and 45$\%$ for $N$=1, never falling below about 40dB signal-to-noise ratio (SNR).
An advantage over an IR-based method is that vibrations are coupled from the interior modules through the exterior surfaces of the robots mechanically, removing any optical property design constraints.

\begin{figure}
    \centering
    \includegraphics[width=\columnwidth]{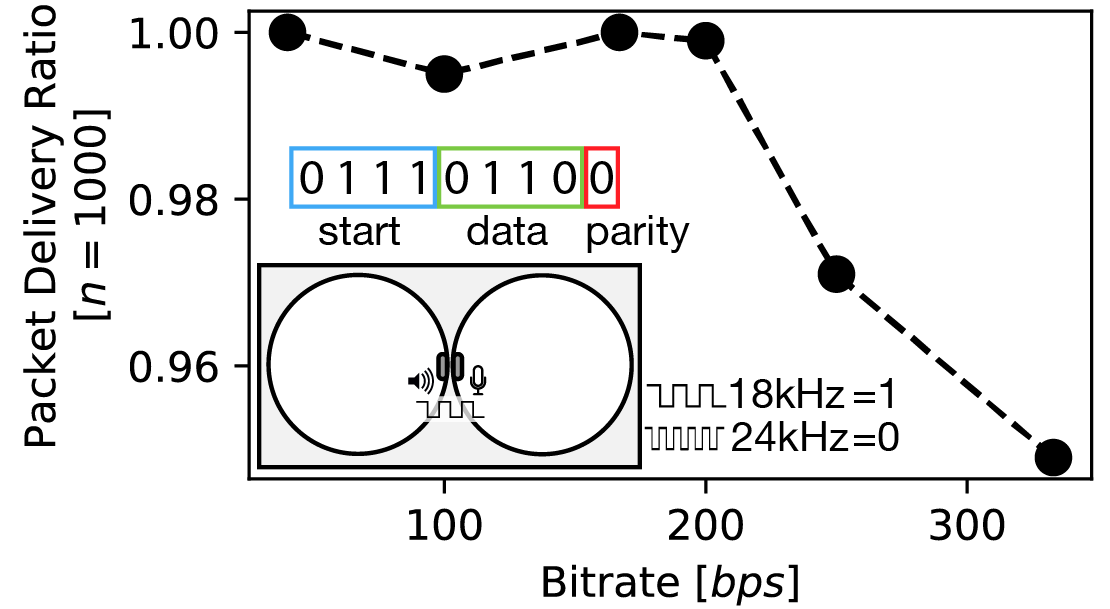}
    \caption{Packet delivery ratio (PDR) versus bitrate in bits/second for a single acoustic module pairing between two robots. Packet delivery ratio is calculated from 1000 packet send attempts, each with randomized data bits. Any bit difference between the sent and received packet is classified as a failure.}
    \label{fig:pdr_datarate}
    \vspace{-4mm}
\end{figure}

We implement acoustic communication through binary frequency-shift keying (FSK), chosen over amplitude modulation in order to resist contact-quality based errors.
A demonstration of the achievable packet delivery ratio (PDR) for a 1:1 module pair is shown in Fig.~\ref{fig:pdr_datarate}. Packets consist of a 4-bit start sequence, 4-bit data structure, and one parity bit. The decrease in achievable PDR is correlated with increasingly tight timing requirements (i.e., a shorter symbol time requires stricter phase alignment) and a decreased SNR caused presumably by the piezoelectric transducers being unable to ring up to full vibration amplitude before a bit transition.

\begin{figure}
    \centering
    \includegraphics[width=.9\columnwidth]{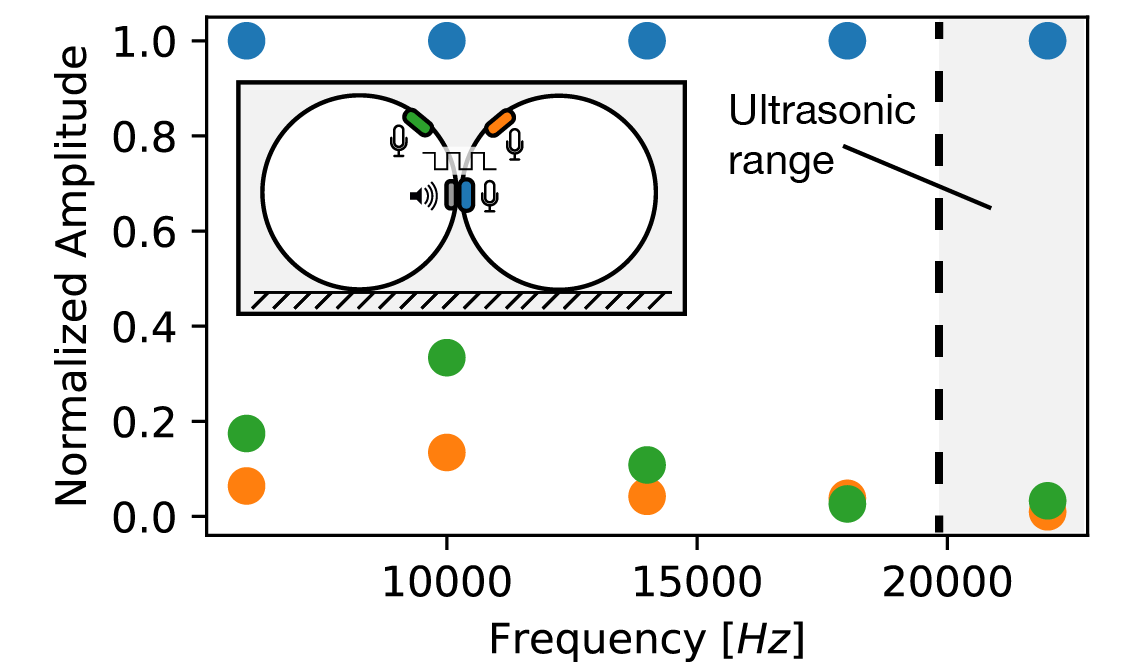}
    \caption{Mean FFT amplitude ($n=20$) at the receiver node and at the neighbors to the receiver and sender, both at approximately 15cm distance, as a function of signal frequency for a pure tone generated by the sending node. Points are normalized to the mean of the amplitude of the receiver node signal for each frequency.}
    \label{fig:crosstalk}
    \vspace{-4mm}
\end{figure}

Having multiple individually addressable communication points on each robot allows for directional communication between any number of connected neighbors.
For this to be possible, signals received at each transducer must be able to be successfully disambiguated from those received at their neighboring nodes; as the signals here are mechanically coupled to the structure and not based on line-of-sight, they radiate symmetrically from their coupling point through the elastic membrane and are received at neighboring points. Fig.~\ref{fig:crosstalk} shows the received signal amplitude at the receiver node versus the received signal amplitude at the neighboring nodes on both the send and receive robots. Vibrations are increasingly attenuated at higher signal frequencies resulting in a higher SNR in the ultrasonic range. 
This means that ultrasonic signals are the best option for sending information directionally through the connections, and can do so with the added benefits of being inaudible and having minimal chance of encountering relevant environmental noise.

For multi-connection data routing over a single channel -- in this case, an individual robot's software-defined FSK receiver, which is only hooked up to a single acoustic module at a time -- we implement a slotless architecture based on the ALOHA protocol~\cite{abramson_aloha_1970}. The default listening behavior is to time multiplex through the $N_{module}$ acoustic modules with an interval equal to a single packet duration $t_{packet}$, waiting to detect a start sequence (\texttt{0111}) and ``locking'' (i.e., remaining listening) if one is detected. If a full packet is decoded with the correct parity bit, an acknowledgement is then sent through the appropriate module. The corresponding sending behavior is to continuously broadcast a packet on all desired output modules for a duration equal to $N_{module} \cdot t_{packet}$, then listen on those modules for the acknowledgement; if no acknowledgement is received the packet is resent.

\subsection{Communication at a Distance}

Collaboration between our robots is possible without either direct contact or line-of-sight via transmission of signals in the audible range, produced effectively by the same piezoelectric transducers thanks to their flat frequency response. In this case, the pressurized elastic skin acts as an omnidirectional pickup for the airborne acoustic waves, letting the ostensibly contact-based piezoelectric transducers act as true microphones. By operating at the approximate resonance of the piezoelectric transducers of 6kHz signals from robots up to a meter away can be received through the air with a measured SNR of $\approx$7dB through the entire operational volume range ($\approx 0.05-0.5m^3$). The received signal amplitude is determined by factors including the robot distance, each robots' volume, and the contact quality between the acoustic modules and the elastic membrane.

One important and fundamental function of decentralized multi-robot systems is the ability to synchronize in time~\cite{trianni2015fundamental}.
In nature, animals use both acoustic and optical (e.g., in katydids~\cite{ravignani2014chorusing} and fireflies~\cite{perez_diaz_firefly-inspired_2016}, respectively) signals to achieve synchronicity in a process known as ``synchronized chorusing,'' or more formally as groups of pulse coupled oscillators.

\begin{figure}
    \centering
    \includegraphics[width=.9\columnwidth]{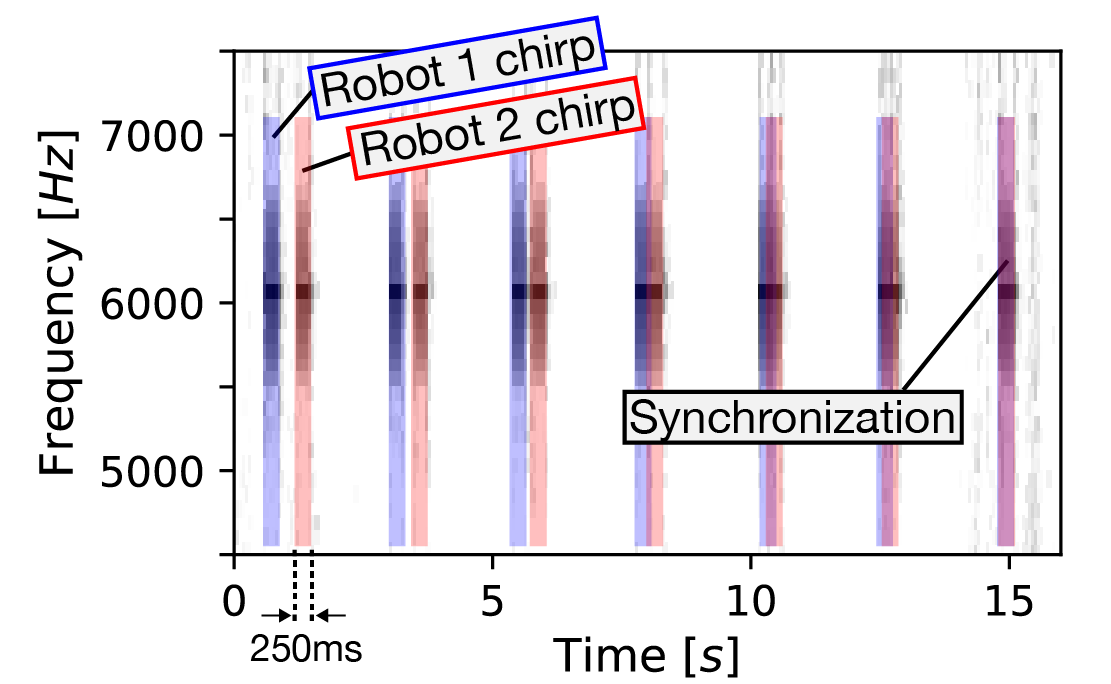}
    \caption{Spectrogram for visual demonstration of clock synchronization between two robots with center-to-center distance of approximately 1m. A randomly determined initial clock gap of approximately 250ms is decreased to less than 5ms after five two-second cycles. For clarity, ambient noise amplitude has been subtracted from the data during post-processing. Data collected using an external microphone.}
    \label{fig:katy_spec}
    \vspace{-4mm}
\end{figure}

Here, pulse coupled synchronization using audible signals is implemented simply; an example spectrogram from synchronization of two robots is shown in Fig.~\ref{fig:katy_spec}. Each robot starts with some initial phase offset from its neighbors (about 250ms in Fig.~\ref{fig:katy_spec}). After a delay $t_{a}$, a synchronization pulse is produced by all $N_{module}$ transducers simultaneously at 6kHz for a duration $t_{chirp}$. The cycle repeats after another delay $t_{b}$. During each delay interval a module acting as the receiver is continuously sampled in order to detect amplitude peaks at 6kHz above a predetermined ambient noise threshold. At the conclusion of the $t_{a}+t_{chirp}+t_{b}$ duration cycle, the tallied detections are used to determine whether the chirp should be shifted ``forward'' or ``backward'' in a binary fashion; if more are detected during $t_{a}$, for example, then the majority of neighboring robots are pulsing \textit{before} this one, so the phase is shifted without changing the period by setting $t_{a} \mathrel{{-}{=}} t_{shift}$ and $t_{b} \mathrel{{+}{=}} t_{shift}$. 

There is a tradeoff between synchronization time and total (audible) robot count. In the most extreme case, all time slots in the listening period would be filled with chirps and therefore balanced.
This means that the time for synchronization is expected to scale with the number of robots as the listening period must increase for additional robots.
Time-varying chirps (such as those produced by katydids~\cite{pipher_frequency_1974}) could provide an additional layer of information that improves the scalability of this approach.

The synchronization accuracy is related to both the chirp duration and the digital signal processing on the receiver. The minimum chirp duration is bounded by the response time of the piezoelectric transducers and the associated SNR at the receiver side. For the receiver processing, non-overlapping 256-point FFT segments with a sampling frequency of 50kS/s results in a minimum synchronization window of approximately 5ms.


\subsection{Exteroceptive Sensing}

Sensing external stimuli like applied loads and environmental contacts is a critical robotic function. 
Existing solutions for soft robots, such as adding flexible signal transmission channels (e.g., optical channels~\cite{xu_optical_2019} or printed traces~\cite{wicaksono_sensornets_2020}) throughout the robot surface, are costly, complex, and not robust to the high-percentage shape change exhibited by our robots.
In order to sense contact we instead take advantage of the coupling between loads on the robot and the resultant attenuation of the acoustic waves being transmitted through the existing unmodified external surface, reducing instrumentation cost and complexity by taking advantage of the compliant nature of the robot.

\begin{figure}
    \centering
    \includegraphics[width=.9\columnwidth]{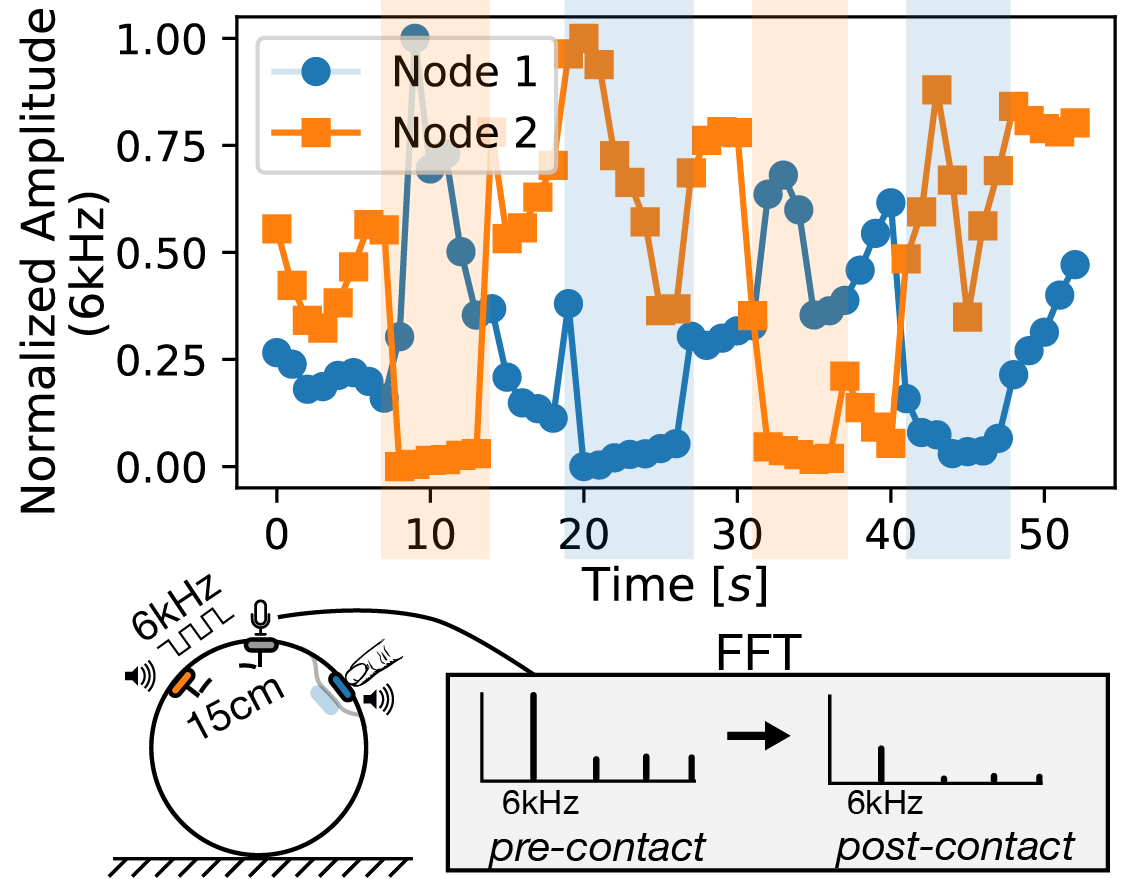}
    \caption{Time-multiplexing the transmission of a pure tone from nodes arranged around a central receiver allows for areas of the robot to be ``sensitized'' to contact. During contact, received amplitude at the central node falls to below 10$\%$ of the initial value. The node emitting the tone switches every 250ms (two alternating nodes shown here) and the FFT results are averaged for a 1Hz update.}
    \label{fig:contact_detection}
    \vspace{-4mm}
\end{figure}

Fig.~\ref{fig:contact_detection} demonstrates that acoustic signals, received at a central receiving node from tones transmitted by surrounding nodes, can be used to detect compression of the robot. Regions are effectively ``sensitized'' by adding a continuously sampling receiver. 
Contacts with areas $\geq a_{mod}$  centered on the transmitting modules both dampen the vibrations of the piezoelectric transducer in its magnetic enclosure as well as decrease the coupling of the surrounding elastic membrane to the node, producing a clearly distinguishable shift in received signal FFT amplitude at the tone frequency.
The sensitive region size is determined by the initial SNR of the received tones, which is a function of inflated volume, pressure, and contact quality. The spatial resolution is determined geometrically by the acoustic module area, $a_{mod}$, the module dispersion density, and the current inflated volume. 
In this inverse to the problem of private contact-based communication, it is important to maximize signal transmission to neighboring nodes and hence requires audible-range signals (see Fig.~\ref{fig:crosstalk}).
Importantly, contact at the receiver node itself manifests as decreases in amplitude from all surrounded nodes; switching the set of ``sensitized'' nodes by reconfiguring the analog crosspoint array could allow for diambiguation.

%% file: 5_behaviors.tex
\section{Autonomous Behavior Demonstrations} \label{behaviors}

\begin{figure*}
    \centering
    \includegraphics[width=\textwidth]{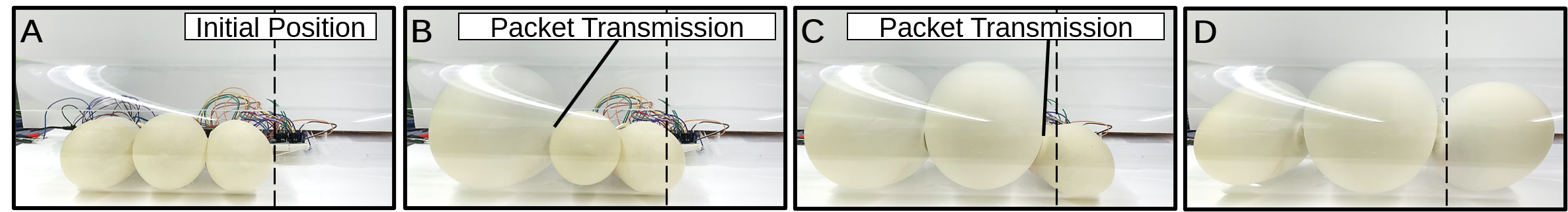}
    \caption{Fully decentralized linear locomotion is possible using the contact-based acoustic communication method developed here. Here, three robots move within a clear cylindrical duct. A) The first robot is commanded to initiate movement. Once it reaches its desired inflation volume (open loop pump control) it sends a data packet through the module on one side of its body. B) The next robot senses a signal at its connector and begins to parse the incoming packet. It understands it is being told to inflate and the cycle continues. C) The signal successfully passes from the first robot to the third robot in the chain. D) If another robot was added to the end before the third had finished inflating it would join in the behavior.}
    \label{fig:locomotion}
\end{figure*}

With 1-DOF actuation, coordination between connected robots allows for locomotion based on an inchworm gait~\cite{devlin_untethered_nodate}. Contact-based communication allows the robots to selectively initiate inflation cycles in neighboring robots. A ``one-dimensional'' locomotion example using this acoustic communication strategy is shown in Figure~\ref{fig:locomotion}. Here, forward motion is only possible when the robots make full contact with the duct walls: the contact detection described in Section~\ref{implementation} could be used to control the inflation and deflation cycles. Locomotion in the X-Y plane could be performed with a minimum group of six such interconnected robots with the ability to communicate in this manner.

\begin{figure*}
    \centering
    \includegraphics[width=.75\textwidth]{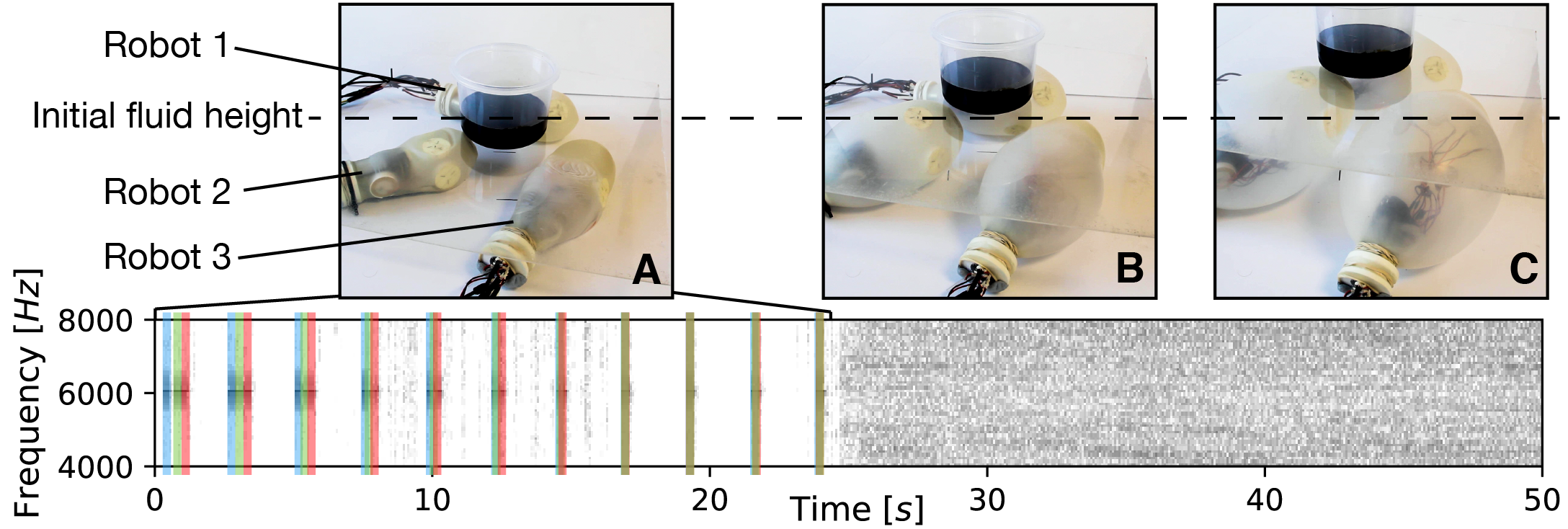}
    \caption{Cooperative lifting of balanced loads is possible via audible-range synchronization at a distance. Here, three robots lift a container of fluid with the load distributed using a sheet of clear acrylic. A) The three robots are commanded to begin a synchronized lift and start communicating through audible chirps with some initial clock skew. B) Once their clocks converge to within a threshold for a set number of periods (four in this experiment) they simultaneously inflate. C) The load is lifted without disturbance. Vertical bars with width roughly equal to chirp duration added to spectrogram for clarity.}
    \label{fig:lifting}
    \vspace{-2mm}
\end{figure*}

Clock synchronization is of practical use for an application like coordinated lifting of unstable or safety-critical objects, such as those theoretically encountered in search and rescue or human-assistive contexts. Figure~\ref{fig:lifting} shows that a group of three robots can lift a balanced load in tandem. A centralized initiation signal tells all three robots to attempt a synchronous lift and sets an initial random clock offset. They begin to use audible-range communication to synchronize (as determined by a maximum number of FFT frames with detected chirps) and once this condition is reached for a minimum of four periods they begin to inflate.

%% file: 6_futurework.tex
\section{Future Work}
There are additional sensing modalities possible using the architecture presented here with relatively minor changes to the system hardware. In the future, sourcing or fabricating properly tuned (i.e., a higher quality factor in the ultrasonic region) or properly coupled (e.g., with an attached acoustic horn) transducers for the acoustic modules may be sufficient for monostatic ultrasonic range finding from each~\cite{borenstein_obstacle_1988}. Bistatic range finding would be an opportunity to take advantage of the shape changing nature of the robots, letting them act as reconfigurable ``acoustic lenses'' which vary field-of-view through changes in volume. 
Contact quality repeatability between modules, and variable contact quality over multiple inflate-deflate cycles, prevented more nuanced force and deformation sensing based on learned models, as in~\cite{laput_sweepsense_2016,swaminathan_input_2019}. A way to more permanently distribute and fix the modules onto the membrane would allow for more functionality. Multi-material composite membranes for the robot exterior could boost SNR through better acoustic impedance matching or add region-dependent sensitivity at design time through acoustic wave guides, as in~\cite{huh_active_2018}.

There are a number of interesting questions related to network architecture for a collection of robots with wide-spectrum transmission capabilities. For example, the audible-range clock synchronization functionality could be used for a slot-based network architecture (e.g., slotted ALOHA), increasing network throughput. In the future, a multi-hop mesh network based on acoustic signals could choose between omnidirectional audible broadcasts and neighbor-to-neighbor ultrasonic modes depending on the traffic route. Additionally, the use of audible range acoustic signals as a primary mode of communication presents opportunities for the study of how human-interpretable modes of multi-robot collaboration affects human operators and bystanders~\cite{moore_making_2017}. 

\section{Conclusion}

Acoustic waves are fundamentally different than electromagnetic (e.g., optical and radio frequency) waves in their transmission properties. Simple and low-cost transducers are available with operation ranges covering broad swaths of the spectrum. By taking advantage of the variable attenuation and directivity of acoustic waves as a function of their frequency, these transducers can be used for functions ranging from communication to sensing. Further, the same high extension ratio pressurized membranes that make soft shape-changing robots difficult to instrument can instead become useful parts of the acoustic transduction strategy by acting as signal channels and state-dependent amplifiers/attenuators.





